\documentclass[]{spie}  

\usepackage{graphicx}
\usepackage{amsmath,amsfonts,amssymb}
\usepackage{subcaption}
\usepackage[colorlinks=true, allcolors=blue]{hyperref}
\usepackage{multirow}
\usepackage{array}
\usepackage{xcolor}
\usepackage[]{xcolor}

\DeclareUnicodeCharacter{03C7}{}

\title{YOLOv8 for Defect Inspection of Hexagonal Directed Self-Assembly Patterns: A Data-Centric Approach}

\author[a,b]{Enrique Dehaerne}
\author[b]{Bappaditya Dey}
\author[c,b]{Hossein Esfandiar}
\author[b]{Lander Verstraete}
\author[b]{Hyo Seon Suh}
\author[b]{Sandip Halder}
\author[a,b]{Stefan De Gendt}
\affil[a]{Faculty of Science, KU Leuven, 3001 Leuven, Belgium}
\affil[b]{Interuniversity Microelectronics Centre (imec), 3001 Leuven, Belgium}
\affil[c]{Dept. of Chemistry, Materials and Chemical Engineering, Giulio Natta; Politecnico di Milano; I-20133 Milano; Italy}

\authorinfo{Send correspondence to Enrique Dehaerne, e-mail: enrique.dehaerne@imec.be}

\begin{document} 
\maketitle

\begin{abstract}
Shrinking pattern dimensions leads to an increased variety of defect types in semiconductor devices. This has spurred innovation in patterning approaches such as Directed self-assembly (DSA) for which no traditional, automatic defect inspection software exists. Machine Learning-based SEM image analysis has become an increasingly popular research topic for defect inspection with supervised ML models often showing the best performance. However, little research has been done on obtaining a dataset with high-quality labels for these supervised models. In this work, we propose a method for obtaining coherent and complete labels for a dataset of hexagonal contact hole DSA patterns while requiring minimal quality control effort from a DSA expert. We show that YOLOv8, a state-of-the-art neural network, achieves defect detection precisions of more than 0.9 mAP on our final dataset which best reflects DSA expert defect labeling expectations. We discuss the strengths and limitations of our proposed labeling approach and suggest directions for future work in data-centric ML-based defect inspection.
\end{abstract}


\section{INTRODUCTION}\label{sect:introduction}

Shrinking semiconductor pattern dimensions has led to increased manufacturing complexity. This complexity has led to a greater number of stochastic defects to inspect. Scanning Electron Microscopy (SEM) is a workhorse fab tool since it can image at high resolution relatively fast. However, SEM images are inherently noisy. Capturing too many frames to reduce this noise can cost time and can be destructive for materials to be inspected such as thin resists. Machine Learning (ML)-based defect detection methods for SEM images are becoming a popular research topic due to their ability to deal with noise better than conventional image processing algorithms\cite{Dey_2022_retinanet}. Additionally, ML-based methods are more easily adaptable to pattern changes (e.g. CD/Pitch variation) and imaging conditions (contrast, field-of-view, etc.). Supervised learning, where an ML model learns by comparing its predictions against (pseudo-)ground truth labels (generally defined by a human expert), usually leads to the best performance. However, the performance of supervised learning depends on the quality of the labels provided by a human labeler. 

In this study, we apply the latest object detection model of the popular YOLO\cite{yolo_og} (You Only Look Once) family, YOLOv8\cite{yolov8_repo}, to detect challenging defects in Hexagonal Directed self-assembly (DSA) patterns. Hexagonal DSA patterns make for an interesting case study as they can contain unique defects that traditional defect inspection software cannot detect or classify. We investigate the effect of variance between human labelers. We consider three labeled datasets, each containing the same SEM images labeled by different human labelers. We propose a data-centric approach to combine labels from multiple labelers and align them with DSA expert expectations. We show that training YOLOv8 models on this final, combined dataset results in detection performance that best matches DSA expert expectations.

\section{Related Work}\label{sect:related_work}
\subsection{Data-centric ML}\label{subsect:data_centric_ml}
Data-centric ML refers to improving an ML system's performance by enhancing the dataset used to train a given model \cite{zha2023datacentric}. This stands in contrast to the model-centric ML approach which, for a given dataset that is assumed to be fixed, makes changes to the ML model to improve the performance of the system. Broadly, methods for data collection/curation \cite{oquab2023dinov2,bogatu2020discovery}, preparation/augmentation \cite{khurana2018feature_engineering,cubuk2019autoaugment}, and labeling \cite{2022rlhf,kirillov2023segment} can all be considered as data-centric improvements for ML systems. 

\subsection{Semiconductor defect inspection}\label{subsect:rw_sem_defect_detection}
Traditionally, semiconductor defect inspection uses hand-crafted image-processing algorithms\cite{auomatic1995breaux,sem2000ritchison,semreview2009McGarvey}. While these algorithms typically perform well for specific inspection scenarios, they struggle to adapt to variations in patterns and imaging conditions. Defect inspection for DSA patterning has been studied using more traditional techniques \cite{dixit2016optical_dsa,ito2014inspection_dsa,pathangi_dsa_2015}. To the best of our knowledge, no ML-based approaches for defect inspection of DSA patterns have been published. ML-based SEM defect detection methods, especially Convolutional Neural Network (CNN) models, are becoming popular due to their improved robustness to these variations. Two main paradigms for training defect inspection models exist, supervised \cite{cheon2019waferdefect,wang2021defectgan,Dey_2022_retinanet,dey2022yolov5} and unsupervised \cite{chang2005selforganizing,Ofir_2022,dey2020pw_estimation} learning. Supervised learning, where an ML model learns by comparing its predictions to labels provided by a human expert, usually leads to the best performance for defect detection and classification. Note, the performance of supervised learning depends on the quality of the labels. The majority of data-centric works in the semiconductor defect inspection domain have focused on data augmentation techniques \cite{dehaerne2023optimizing,naoaki2018wacv}. In this study, we propose improving label quality as another method for improving the performance of CNN-based semiconductor defect detection systems.

\section{Hexagonal Directed Self-Assembly SEM image Dataset}\label{sect:dataset}
For this research work, we collected a novel dataset of hexagonal DSA SEM images with defects. DSA is a patterning technique that uses the inherent period of block-copolymer (BCP) materials to define line/space or hexagonal hole patterns from the bottom up \cite{paulina_dsa_2012,arjun_dsa_2016}. To generate well-aligned patterns, a guiding pattern for the BCP materials is needed. While this guiding pattern is typically created by conventional lithography techniques, the final pattern characteristics are defined by the self-assembled BCP materials. This allows DSA to overcome stochastic issues commonly associated with EUV lithography. However, DSA is known to have specific defects that are otherwise not observed in conventional lithography patterns. These DSA specific defects are associated with the dynamic processes underlaying the self-assembly process. Typical DSA specific defects are dislocations in line/space patterns, which represent a kinetically trapped metastable structure \cite{pathangi_dsa_2015,doise_dsa_2019}. Similarly, HEXagonal Contact Hole (HEXCH) arrays may suffer from misaligned domains and even non-vertically oriented structures. When etching the latter into the underlying hard mask, closed holes may be observed.

\begin{figure}
\centering
    \begin{subfigure}{0.24\linewidth}
\includegraphics[width=\linewidth]{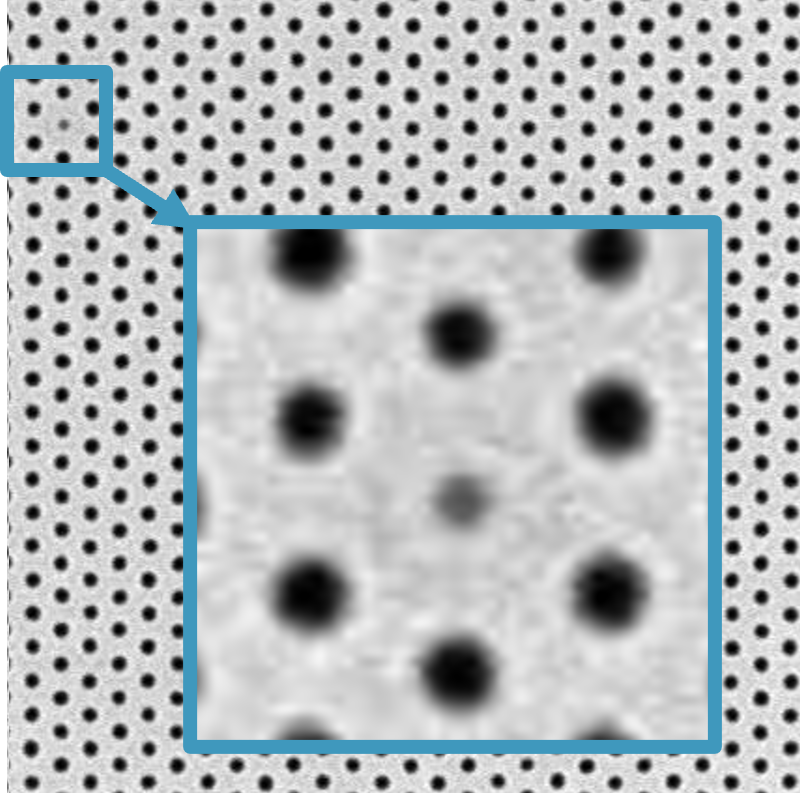} 
    \caption{Partially Closed Hole (PCH)}
    \end{subfigure}\hfill
    \begin{subfigure}{0.24\linewidth}
\includegraphics[width=\linewidth]{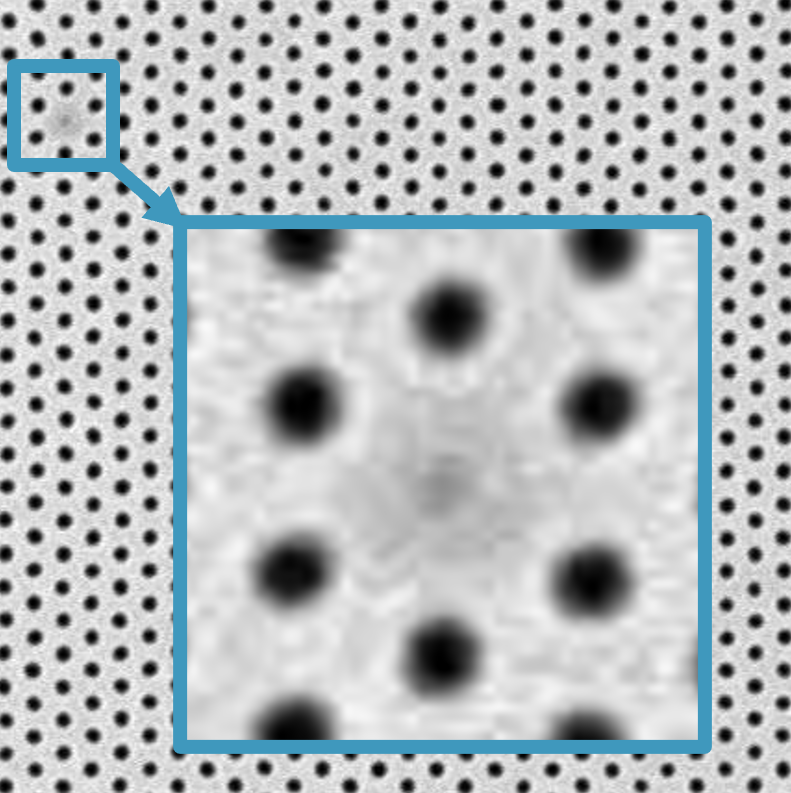}
    \caption{Missing Hole (MH)\newline}
    \end{subfigure}\hfill
    \begin{subfigure}{0.24\linewidth}
\includegraphics[width=\linewidth]{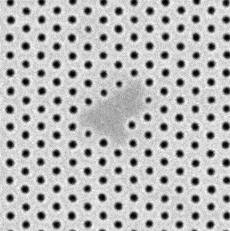} 
    \caption{Closed Patch (CP)\newline}
    \end{subfigure}\hfill
    \begin{subfigure}{0.24\linewidth}
\includegraphics[width=\linewidth]{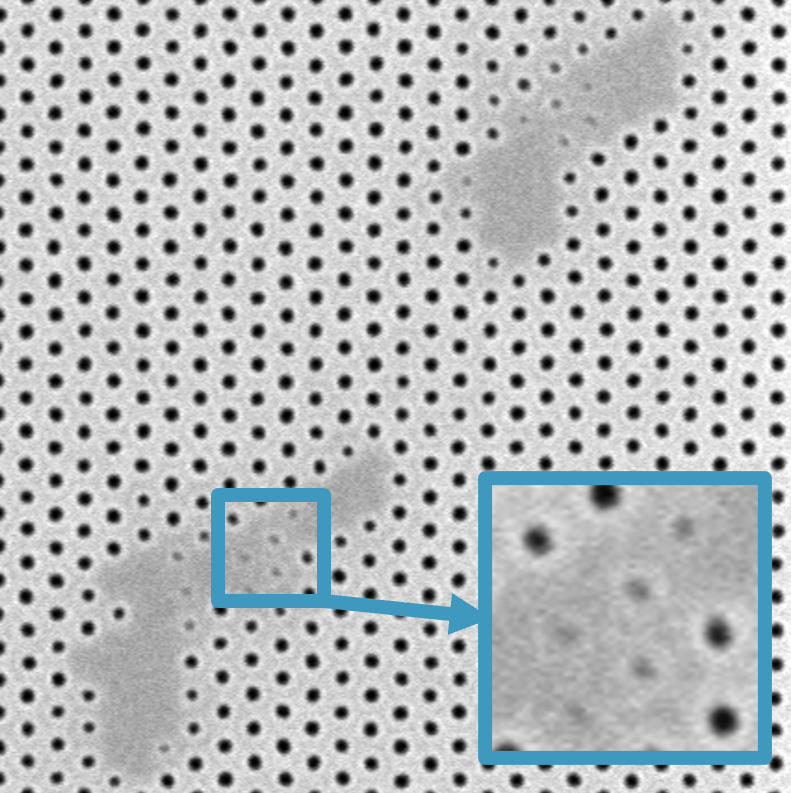} 
    \caption{Multiple Defect (MD)\newline}
    \label{subfig:multiple_defect_example}
    \end{subfigure}\hfill
\caption{Defect examples for each raw image partition of the full hexagonal DSA dataset. The images are cropped (and zoomed in where necessary).}
    \label{fig:example_crops}
\end{figure}

Our novel dataset consists of SEM images of HEXCH DSA pattern defects after etch into a SiN layer. In these images, we localize (in the form of bounding boxes) and classify three different classes of defects. These defect classes are Partially Closed Holes (PCH), Missing Holes (MH), and Closed Patches (CP). PCH and MH defects each relate to one hole while a CP defect relates to a contiguous area of closed holes. As DSA HEXCH patterns are not yet routinely considered in the semiconductor industry, defects in these patterns are not covered in traditional inspection methodology. The DSA dataset images were collected by a DSA patterning expert. The expert inspected each image and partitioned the collection into four sub-collections based on the defects shown in each image. The names given to these partitions correspond to the three defect classes listed above (PCH, MH, and CP) as well as a Multiple Defect (MD) sub-collection. These MD images contained more than one defect of one or more different defect classes. Figure \ref{fig:example_crops} shows examples of defects of these four partitions. The full dataset organized into these four sub-collections was given to three different labelers with instructions to draw and classify bounding boxes that best capture the extent of each defect in each image. In the rest of this paper we refer to the labelers as \textit{labeler A}, \textit{labeler B}, and \textit{labeler C}. Note that this initial partitioning of the dataset is not necessary for the labeling methodology proposed in Section \ref{sect:methodology}.

The dataset was randomly split into train, validation, and test subsets with a ratio of 70:15:15, respectively. The image-level statistics of the dataset are provided in Table \ref{tab:raw_dataset}. Each SEM image has a resolution of 1000$\times$1000 pixels.

\begin{table}[h]
    \caption{SEM image counts for each split and partition of the HEXCH DSA dataset.}
    \label{tab:raw_dataset}
    \begin{center}
    \begin{tabular}{|c|c|c|c|c|}
        \hline
        \textbf{Partition} & \textbf{Train} & \textbf{Validation} & \textbf{Test} & \textbf{All} \\ 
        \hline
        Partially Closed Hole (PCH) & 73 & 13 & 18 & 104 \\ 
        Missing Hole (MH) & 29 & 8 & 5 & 42 \\
        Closed Patch (CP) & 72 & 13 & 19 & 104 \\
        Multiple Defect (MD) & 63 & 17 & 9 & 89 \\ 
        \hline
        Total & 237 & 51 & 51 & 339 \\ 
        \hline
    \end{tabular}
    \end{center}
\end{table}

\section{Methodology}\label{sect:methodology}

\subsection{Semi-automated labeler consensus}\label{subsect:labeler_consensus}

To mitigate individual labeler errors and biases, we propose a semi-automated method for combining labels from multiple labelers. A naive method for combining multiple labeled datasets would be to have a DSA expert manually compare labels for each image and remove or modify each label individually. This naive method is inefficient for two main reasons. First, a majority of labels will most likely have consensus in the sense that many labels between multiple labelers overlap to a high degree and are classified as the same class. Second, misunderstanding labeling instructions may lead to common patterns of erroneous labels. Oftentimes, these common patterns can be easily identifiable and automatically corrected without requiring a DSA expert's manual intervention. To mitigate these inefficiencies, respectively, our proposed method includes: (i) using voting for reclassification and Weighted-Box-Fusion (WBF) \cite{solovyev2021weighted} for bounding box extent to combine overlapping labels and (ii) programmatically correcting common labeling error patterns as identified by a DSA expert. In this way, the proposed method only requires DSA expert intervention two times: (a) breaking ties between labelers during reclassification voting and (b) identifying common erroneous label patterns.


\begin{figure}
\centering
\includegraphics[width=\linewidth]{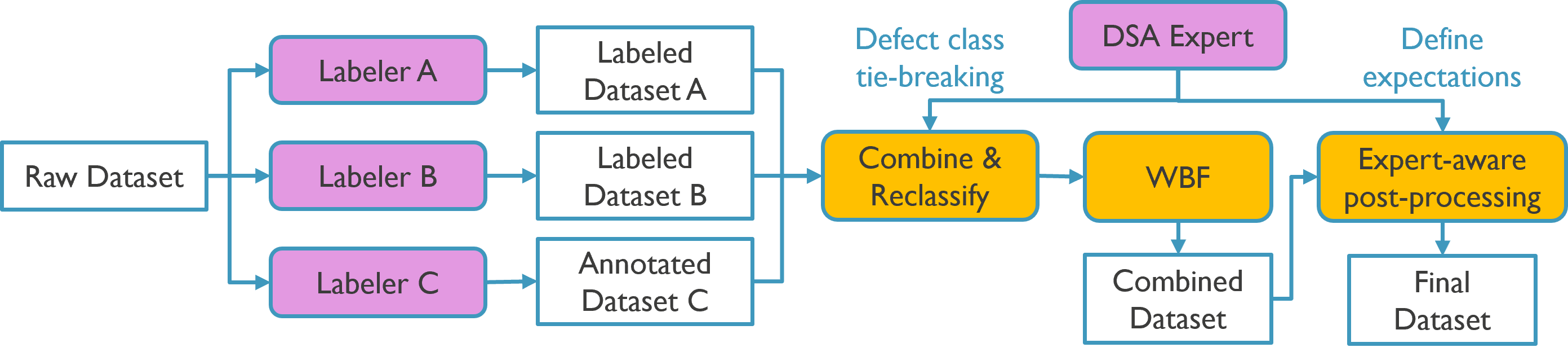}
\caption{Flowchart of the proposed labeling and consensus methodology.}
    \label{fig:consensus_algo_diagram}
\end{figure}

Figure \ref{fig:consensus_algo_diagram} shows a flowchart of the proposed labeling procedure. First, the raw SEM image dataset is labeled by multiple labelers (three labelers in the case of this study) to obtain multiple labeled datasets. Note that this part does not require any consensus between the labelers which means each labeler can work independently and in parallel, optimizing labeling speed. The labels of all the labelers are then combined on all corresponding images. Standard labeling software was used by the labelers. LabelImg \cite{labelimg} was used by labelers B and C. Coco-annotator \cite{cocoannotator} was used by labeler A. 

Second, overlapping clusters of labels are reclassified based on voting between these labels. A cluster of labels is formed by selecting a label and adding all other labels in the image that have an Intersection-over-Union (IoU) value greater or equal to $0.5$ to a cluster along with the selected label. This is similar to how the WBF algorithm clusters labels to be fused together.  Within a cluster, a vote is performed using each label's class. All labels in the cluster are then reclassified to the class with the most votes. Reclassification vote ties are resolved manually by a DSA expert.

\begin{figure}
    \centering
    \begin{subfigure}{0.24\linewidth}
\includegraphics[width=\linewidth]{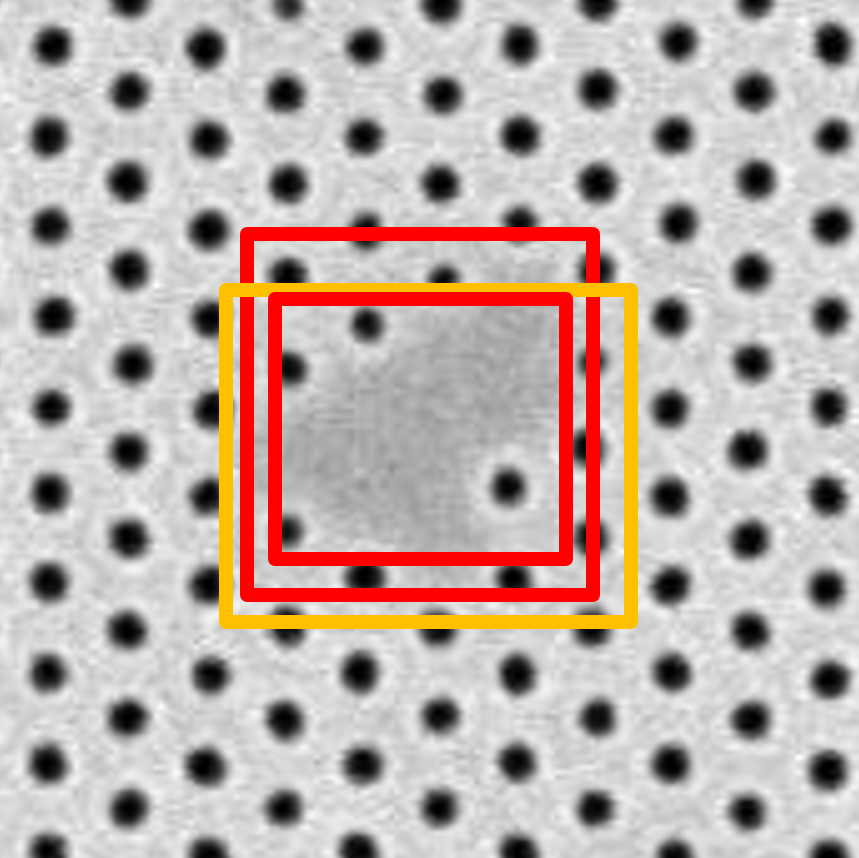}
    \caption{Example labels for a defect.}
    \label{subfig:label_example_a}
    \end{subfigure}\hfill
    \begin{subfigure}{0.26\linewidth}
\includegraphics[trim={0 15pt 0 0},width=\linewidth]{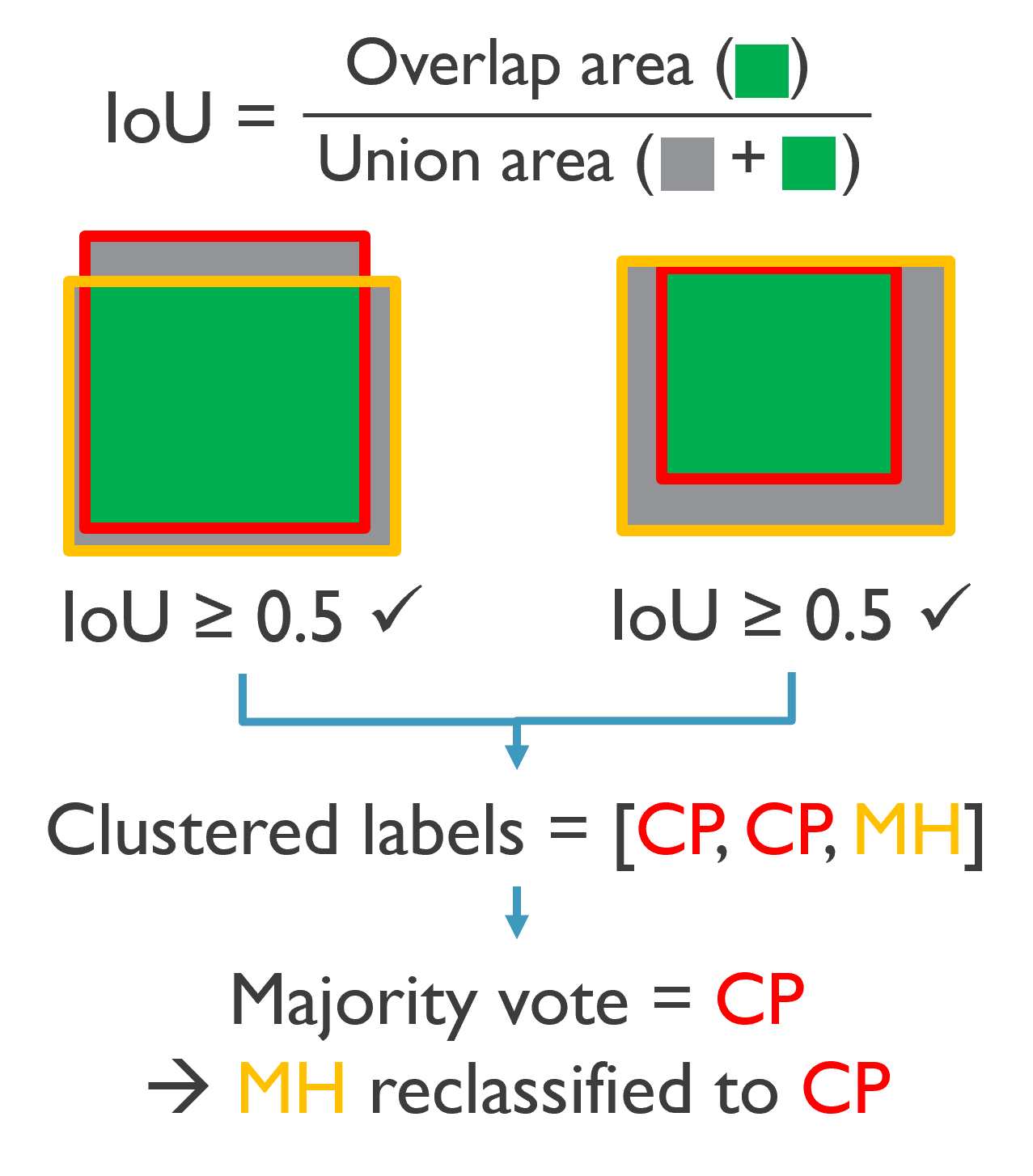}
    \caption{IoU calculation for overlapping labels.}
    \label{subfig:label_example_b}
    \end{subfigure}\hfill
    \begin{subfigure}{0.24\linewidth}
\includegraphics[width=\linewidth]{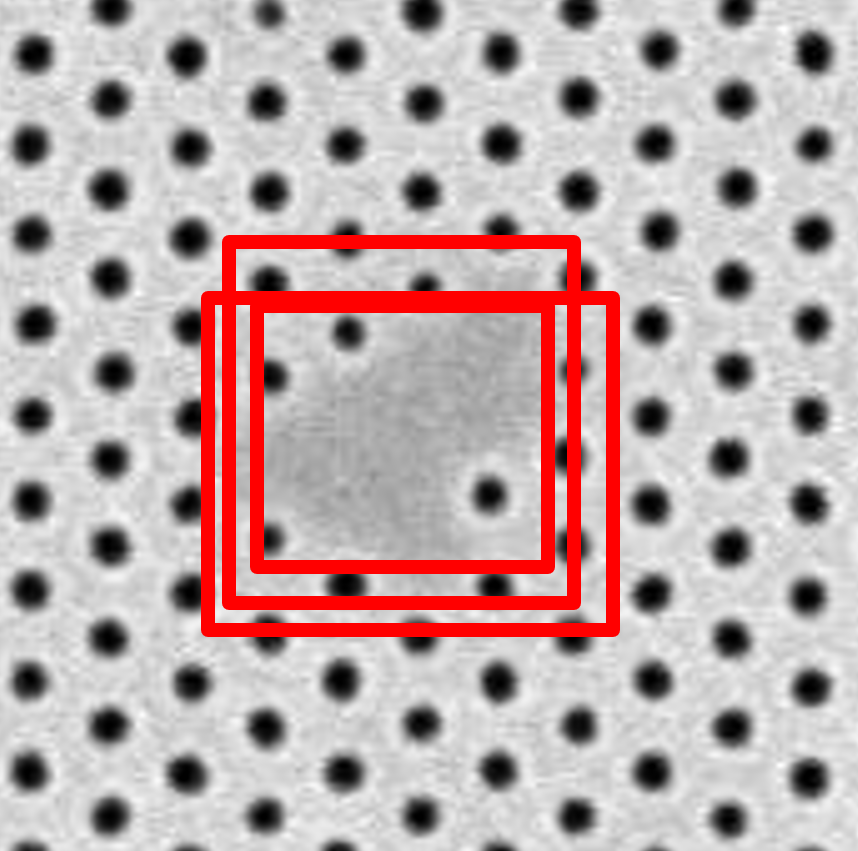}
    \caption{IoU calculation for overlapping labels.}
    \label{subfig:label_example_c}
    \end{subfigure}\hfill
    \begin{subfigure}{0.24\linewidth}
\includegraphics[trim={0 0 0 10pt}, width=\linewidth]{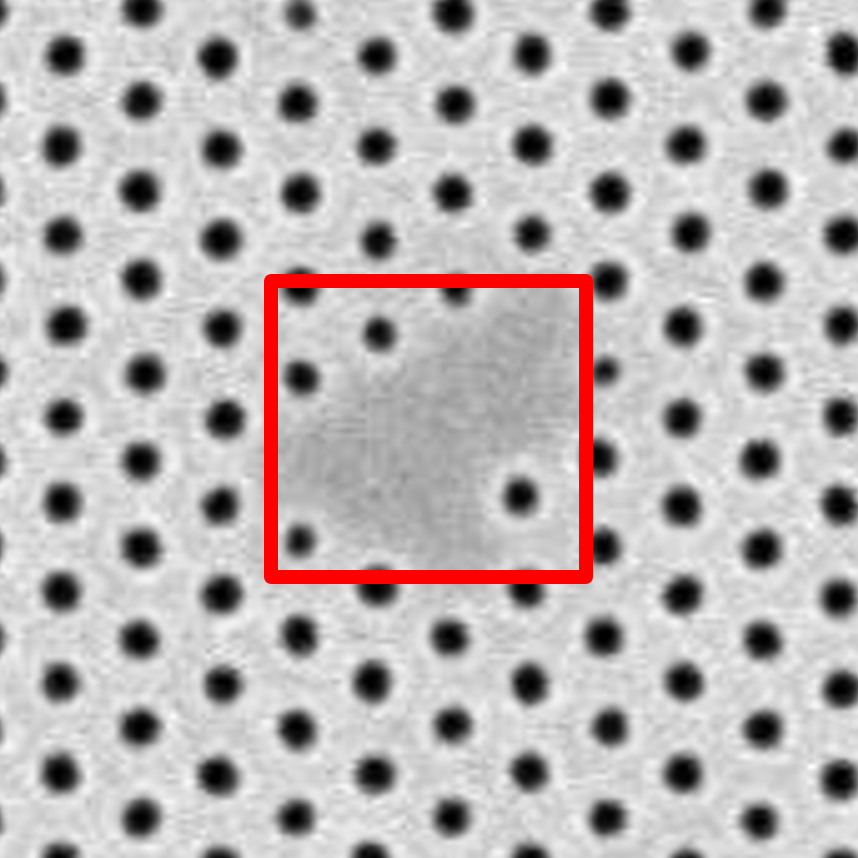}
    \caption{Example labels after WBF.}
    \label{subfig:label_example_d}
    \end{subfigure}
    \caption{An example scenario with three overlapping labels, two \textcolor{red}{CP} labels and one \textcolor{orange}{MH} label, (\ref{subfig:label_example_a}). Using the orange label as a randomly selected starting label, the labels are clustered together based on having IoU values greater or equal to $0.5$ (\ref{subfig:label_example_b}). The most common class in this label cluster is \textcolor{red}{CP} so the \textcolor{orange}{MH} label is changed to a \textcolor{red}{CP} label (\ref{subfig:label_example_b} and \ref{subfig:label_example_c}). The cluster of labels is then fused into a single label using the WBF algorithm (\ref{subfig:label_example_d}).}
    \label{fig:iou_wbf_example}
\end{figure}

The (possibly reclassified) overlapping clusters of labels are fused into one label via WBF. The WBF algorithm expects as input a list of bounding boxes, each with corresponding labels and confidence scores. Compared to other bounding-box filtering methods such as Non-Maximum Suppression (NMS), WBF does not rely on bounding-boxes having varying confidence scores. This means that we can assign the same confidence score, $1$, to all labels since we do not know the confidence of a label. Additionally, WBF has the desirable property that it forms a new bounding box using properties from all boxes in a cluster rather than simply throwing away boxes such as with NMS. After WBF is applied, a \textit{combined dataset} is obtained which can more easily be inspected by a DSA expert than the raw combined labels of all labelers. Figure \ref{fig:iou_wbf_example} shows an example of how labels are analyzed for overlap, reclassified via voting, and fused using WBF. The combined dataset is obtained after WBF has been applied to all images of the dataset results in the \textit{combined} dataset.

While it is possible to let the DSA expert thoroughly inspect and correct each label instance of the \textit{combined} dataset, we find that a majority of the labeling errors follow a pattern that the DSA expert can identify by only inspecting a few labeled image samples from the combined dataset. We also find that these common label error patterns can be corrected using automatic post-processing, mitigating the need for the DSA expert to correct these errors manually. The exact required post-processing steps will be different for different labelers and different datasets. The post-processing steps used in our study are discussed in Section \ref{subsection:results_labeler_diff}. If very few errors are found, it will most likely be faster to correct these individually. As with all steps in the proposed methodology, using post-processing for error correction scales better with dataset size and erroneous label counts compared to manual label correction.

Following the methodology described above, three datasets labeled by individual labelers (\textit{labelers} \textit{A}, \textit{B}, and \textit{C}) and two labeled datasets from combining and processing these three datasets (\textit{combined} and \textit{final}, respectively) were obtained. The final dataset is considered to be the labeled dataset most similar to the DSA expert's ideal dataset. The defect detection performance results on the final dataset are therefore considered to be the most important.

\subsection{YOLOv8 defect detection model}\label{subsect:yolov8}
The original YOLO model \cite{yolo_og} was developed to achieve a good trade-off between detection performance and inference speed. This is a very desirable characteristic for industrial image-based defect inspection applications where throughput is important. YOLOv8 \cite{yolov8_repo} is a recent YOLO-style, single-stage object detection model that has been shown to outperform previous YOLO models in terms of mean Average Precision (mAP) on the COCO dataset \cite{coco}.
YOLOv8 comes in five different size variants: \textit{nano}, \textit{small}, \textit{medium}, \textit{large}, and \textit{extra-large}. In our experiments, we train, validate, and test each variant on the various labeled datasets obtained from the labeling methodology proposed in the previous subsection. We report the average test results over all of the variants to obtain our final AP results. No consistent differences between model size variants were observed for the experiments conducted.

All experiments were run on a Tesla V100 GPU with 32GB of VRAM unless mentioned otherwise. Images were resized to the closest multiple of 32, 992$\times$992, since YOLOv8 models have a maximum stride of 32. A batch size of 32 was used for the \textit{nano}, \textit{small}, and \textit{medium} YOLOv8 model variants. A batch size of 16 was used for the \textit{large} and \textit{extra-large} variants. All models were trained for a maximum of 200 epochs. The checkpoint that achieves the best fitness value ($fitness = 0.1\times mAP@0.5 + 0.9\times mAP@0.5:0.95$), as calculated at the end of each epoch, was used for evaluation. The training for all experiments was automatically stopped if the fitness value did not improve after 50 epochs. All other hyperparameters were kept the same as the default settings of the YOLOv8 GitHub repository \cite{yolov8_repo}. Most training runs were completed within 2 hours.

\section{Results \& Discussion}\label{sect:results}

\subsection{Differences in labeled datasets}\label{subsection:results_labeler_diff}

\begin{figure}
\centering
    \begin{subfigure}{0.24\linewidth}
    \caption*{Original (cropped) images}
\includegraphics[width=\linewidth]{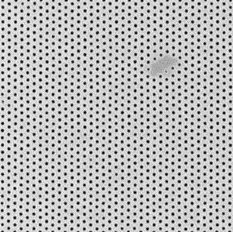}
    \end{subfigure}\hfill
    \begin{subfigure}{0.24\linewidth}
    \caption*{Labels from labeler A}
\includegraphics[width=\linewidth]{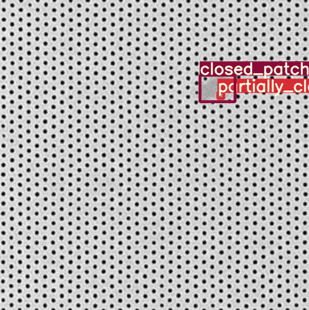}
    \end{subfigure}\hfill
    \begin{subfigure}{0.24\linewidth}
    \caption*{Labels from labeler B}
\includegraphics[width=\linewidth]{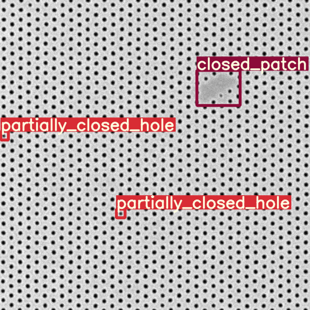}
    \end{subfigure}\hfill
    \begin{subfigure}{0.24\linewidth}
    \caption*{Labels from labeler C}
\includegraphics[width=\linewidth]{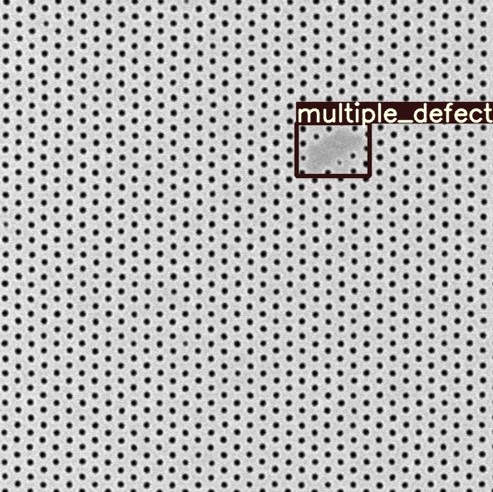}
    \end{subfigure}\hfill
    \begin{subfigure}{0.24\linewidth}
\includegraphics[width=\linewidth]{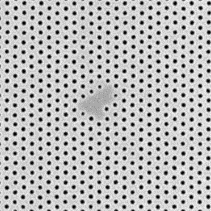}
    \end{subfigure}\hfill
    \begin{subfigure}{0.24\linewidth}
\includegraphics[width=\linewidth]{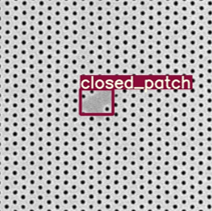}
    \end{subfigure}\hfill
    \begin{subfigure}{0.24\linewidth}
\includegraphics[width=\linewidth]{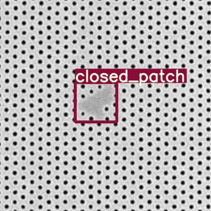}
    \end{subfigure}\hfill
    \begin{subfigure}{0.24\linewidth}
\includegraphics[width=\linewidth]{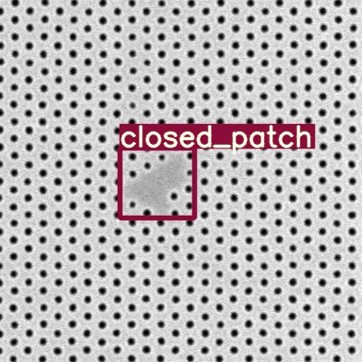}
    \end{subfigure}\hfill
    \begin{subfigure}{0.24\linewidth}
\includegraphics[width=\linewidth]{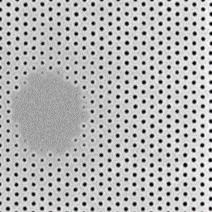} 
    \end{subfigure}\hfill
    \begin{subfigure}{0.24\linewidth}
\includegraphics[width=\linewidth]{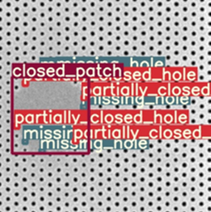}
    \end{subfigure}\hfill
    \begin{subfigure}{0.24\linewidth}
\includegraphics[width=\linewidth]{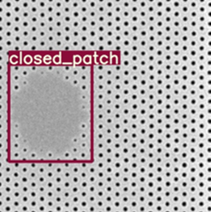} 
    \end{subfigure}\hfill
    \begin{subfigure}{0.24\linewidth}
\includegraphics[width=\linewidth]{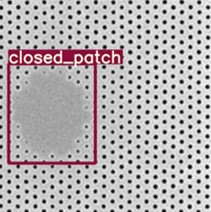} 
    \end{subfigure}\hfill
    \begin{subfigure}{0.24\linewidth}
\includegraphics[width=\linewidth]{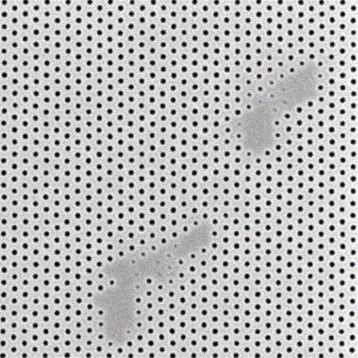}
    \end{subfigure}\hfill
    \begin{subfigure}{0.24\linewidth}
\includegraphics[width=\linewidth]{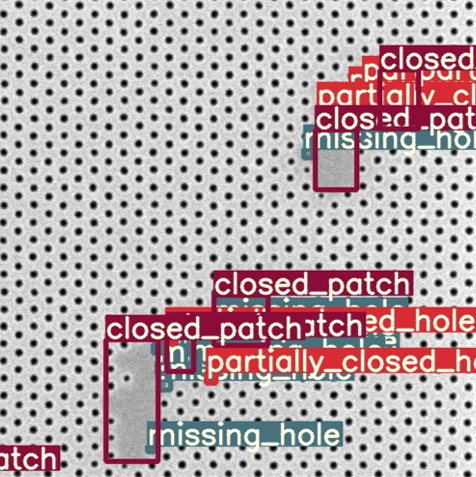}
    \end{subfigure}\hfill
    \begin{subfigure}{0.24\linewidth}
\includegraphics[width=\linewidth]{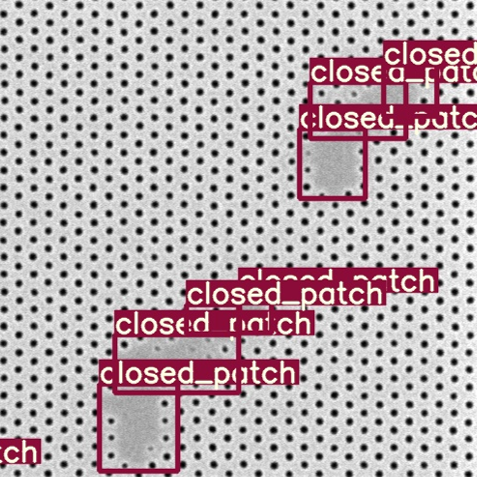}
    \end{subfigure}\hfill
    \begin{subfigure}{0.24\linewidth}
\includegraphics[width=\linewidth]{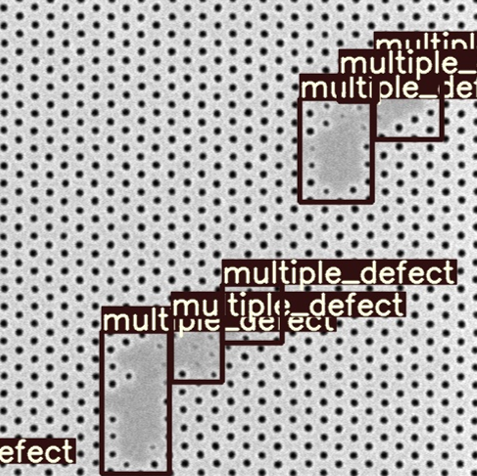}
    \end{subfigure}\hfill
\caption{Examples of labeling differences between labelers A, B, and C.}
    \label{fig:label_difference_examples}
\end{figure}

Figure \ref{fig:label_difference_examples} shows examples of prominent labeling behavior differences between the three labelers. The labeling differences in each row and how these differences were handled are described below: 
\begin{enumerate}
    \item The first row shows labeler A missing or not considering certain holes to be partially closed when labeler B does. Combining labels from all the labelers ensures that as few defects as possible are missed in the final labeled dataset. The first row also shows labeler C wrongly classifying all defects in the MD partition as a separate MD defect class. This is corrected through class voting. For example, in this particular instance, the label from labeler C would be reclassified as a CP since it is the majority vote when considering the overlapping labels from the three labelers.
    \item The exact extent of closed patches bounding boxes is different between the three labels in the second row. This is solved by using WBF to fuse the labels into a single label of the same class that has an extent somewhere between all the overlapping labels.
    \item The third row shows labelers B and C did not consider labeling partially closed and missing holes instances inside of a closed patch instance separately, while labeler A labeled all of these instances separately. While the image features captured by labeler A's labels may very well reflect true MH or PCH defect instances, the DSA expert preferred B and C's labeling. Therefore, for the final dataset, all combined image labels were processed to remove MH and PCH labels within CP labels. 
    \item The last row again shows some of the labeling differences from the previous row but also shows irregularly shaped closed patches that were broken up into multiple bounding boxes by each labeler differently. While these multiple boxes more precisely encapsulate CP features without capturing non-defective background features, the DSA expert preferred them to be labeled as one CP instead of multiple, separate CPs. For the final dataset, combined image labels were processed to combine overlapping CPs (in this case using an IoU threshold of more than zero) into the smallest box that completely covers all overlapping CPs labels. For this post-processing step, we use an IoU threshold of $0.0001$ since most of these adjacent CP boxes barely overlap.
\end{enumerate}

\begin{figure}
    \centering
        \begin{subfigure}{0.45\linewidth}
    \includegraphics[width=\linewidth]{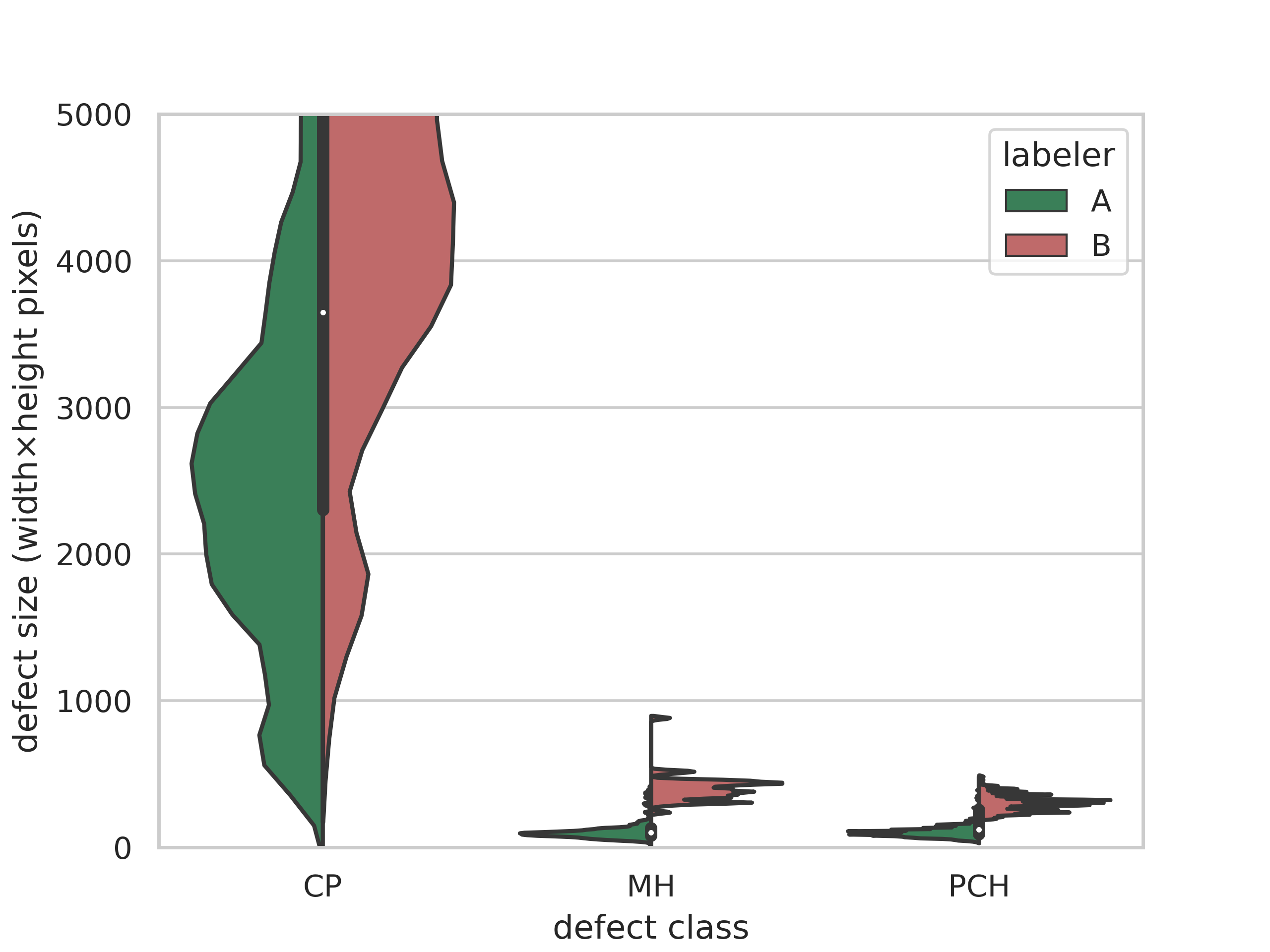} 
    \caption{Labelers A and B}
        \end{subfigure}
        \begin{subfigure}{0.45\linewidth}
    \includegraphics[width=\linewidth]{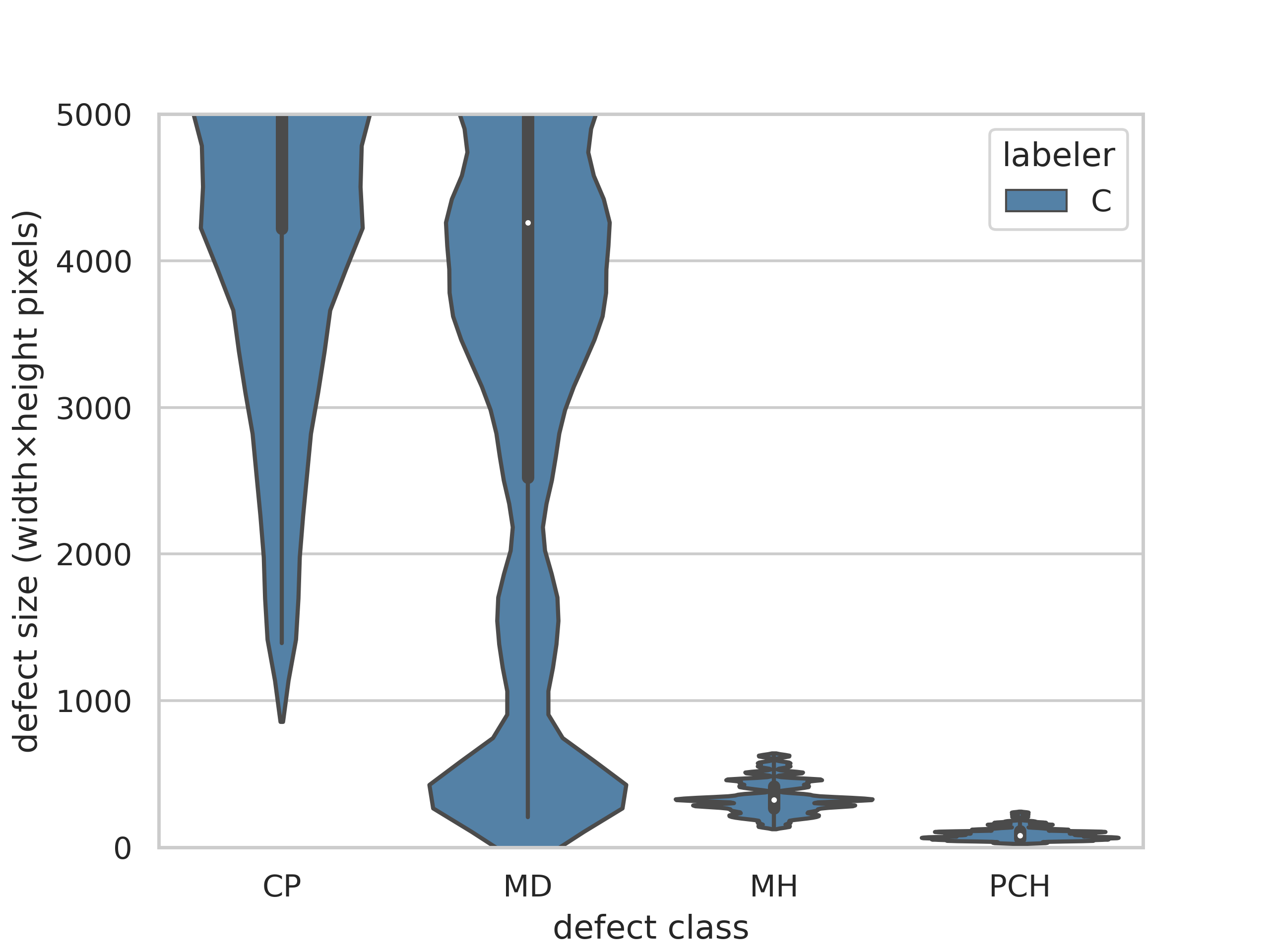} 
    \caption{Labeler C}
        \end{subfigure}
\caption{Violin plots for the defect size of labels of the three individually labeled datasets. Each violin is scaled to have the same width. The plots are cut off at a defect size of 5000 to improve readability.}
\label{fig:label_violin_plots}
\end{figure}

After combining and labeling class voting, MD labels that did not overlap with labels from labelers A or B still remained. As can be seen in Figure 
\ref{fig:label_violin_plots}, the sizes of CP, MH, and PCH labels for labeler C were particularly well separable. Therefore, the remaining MD labels were reclassified to be CPs if their size was larger than the largest MH or PCH label from labelers C.  Otherwise, the remaining MDs were reclassified to be PCHs. This was done because the sizes of the remaining MD labels were much more similar to the average size of labeler C's PCH labels. It can be assumed that PCH labels are more easily to be accidentally missed by labelers than MH labels which further motivated the choice to reclassify the remaining MDs as PCHs instead of MHs.

\begin{figure}
\centering
    \begin{subfigure}{0.3\linewidth}
    \caption*{Original (cropped) images}
\includegraphics[width=\linewidth]{abstract_images/nolabel_md_Site_41_cropped.png}
    \end{subfigure}\hfill
    \begin{subfigure}{0.3\linewidth}
    \caption*{Labels from combined dataset}
\includegraphics[width=\linewidth]{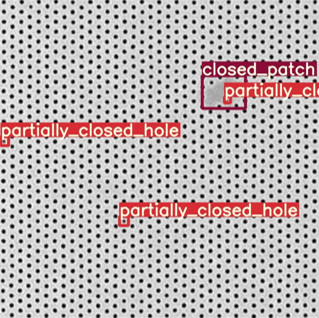}
    \end{subfigure}\hfill
    \begin{subfigure}{0.3\linewidth}
    \caption*{Labels from final dataset}
\includegraphics[width=\linewidth]{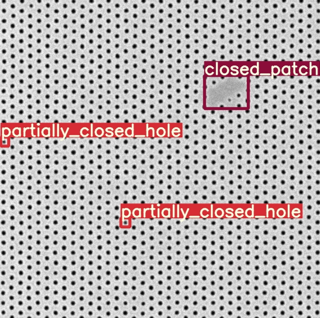}
    \end{subfigure}\hfill
    \begin{subfigure}{0.3\linewidth}
\includegraphics[width=\linewidth]{abstract_images/nolabel_cp_Site_74_cropped.png}
    \end{subfigure}\hfill
    \begin{subfigure}{0.3\linewidth}
\includegraphics[width=\linewidth]{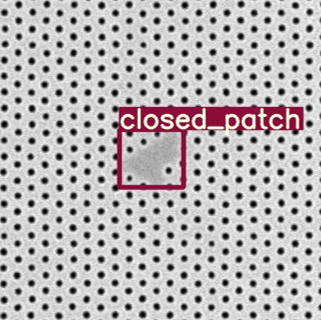}
    \end{subfigure}\hfill
    \begin{subfigure}{0.3\linewidth}
\includegraphics[width=\linewidth]{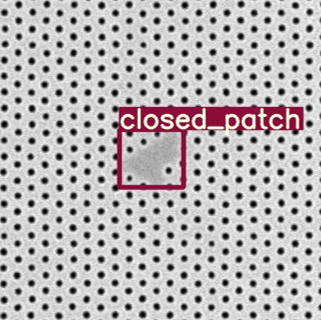}
    \end{subfigure}\hfill
    \begin{subfigure}{0.3\linewidth}
\includegraphics[width=\linewidth]{abstract_images/nolabel_cp_Site_5956_cropped.png} 
    \end{subfigure}\hfill
    \begin{subfigure}{0.3\linewidth}
\includegraphics[width=\linewidth]{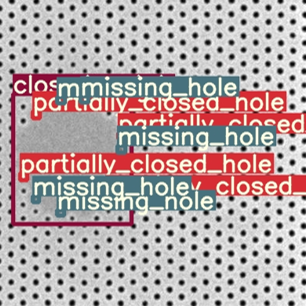}
    \end{subfigure}\hfill
    \begin{subfigure}{0.3\linewidth}
\includegraphics[width=\linewidth]{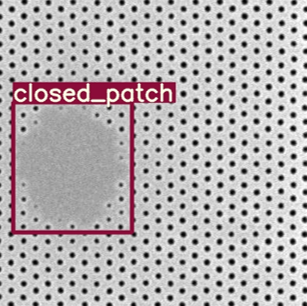} 
    \end{subfigure}\hfill
    \begin{subfigure}{0.3\linewidth}
\includegraphics[width=\linewidth]{images/nolabel_md_site_227_cropped.png}
    \end{subfigure}\hfill
    \begin{subfigure}{0.3\linewidth}
\includegraphics[width=\linewidth]{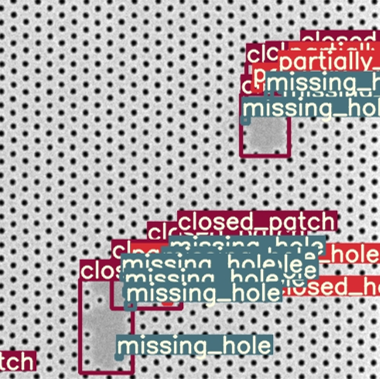}
    \end{subfigure}\hfill
    \begin{subfigure}{0.3\linewidth}
\includegraphics[width=\linewidth]{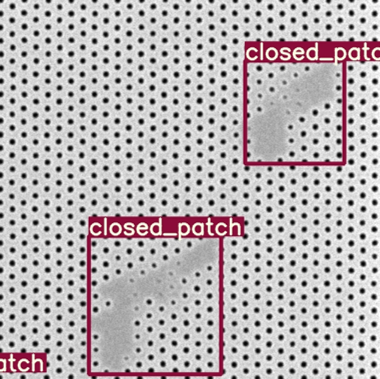}
    \end{subfigure}\hfill
\caption{Examples of labeling differences between the combined and final datasets.}
    \label{fig:consensus_label_difference_examples}
\end{figure}

\begin{figure}
    \centering
        \begin{subfigure}{0.45\linewidth}
    \includegraphics[width=\linewidth]{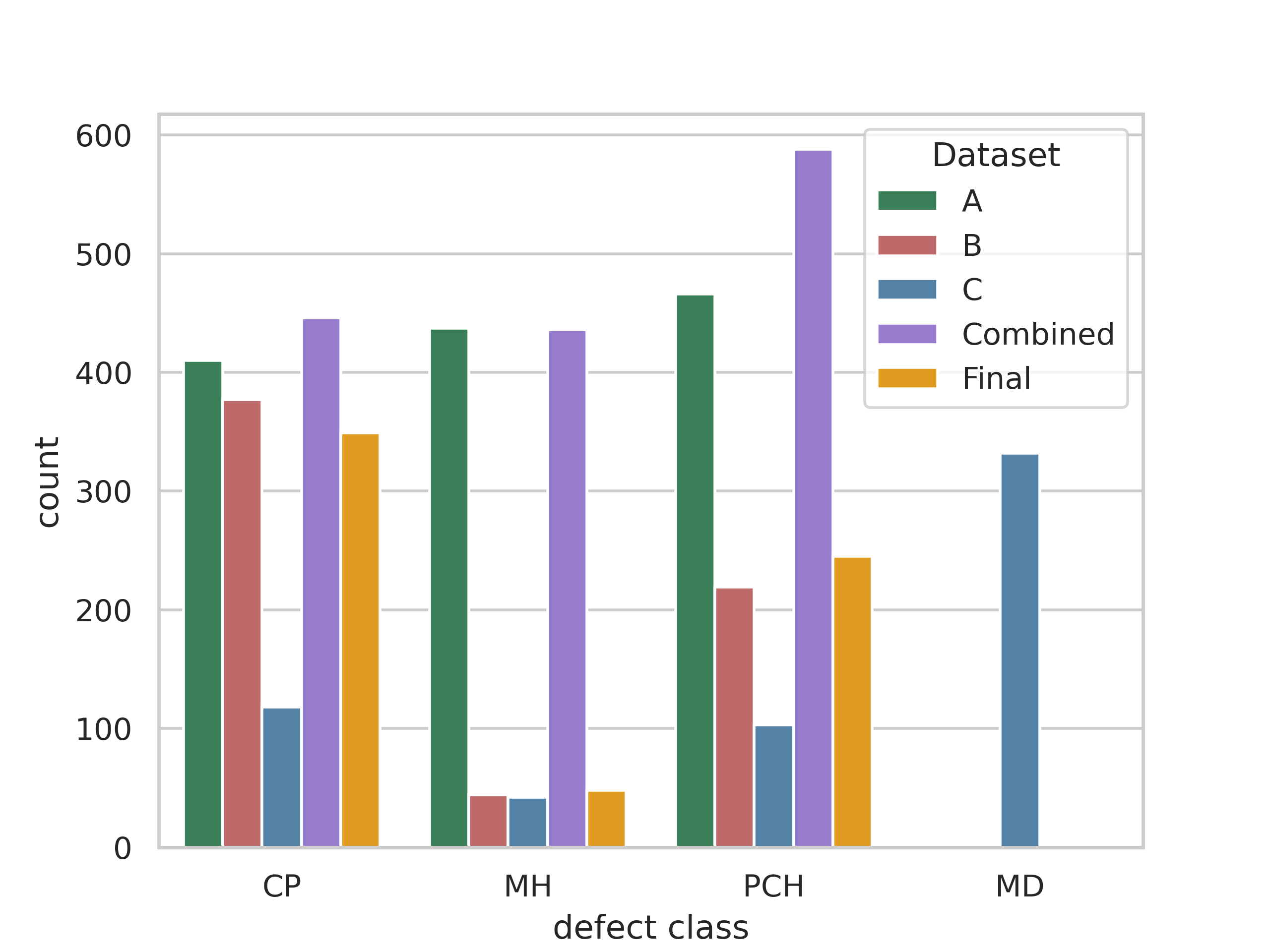} 
    \caption{All images}
    \label{subfig:label_bar_graphs_all}
        \end{subfigure}
        \begin{subfigure}{0.45\linewidth}
    \includegraphics[width=\linewidth]{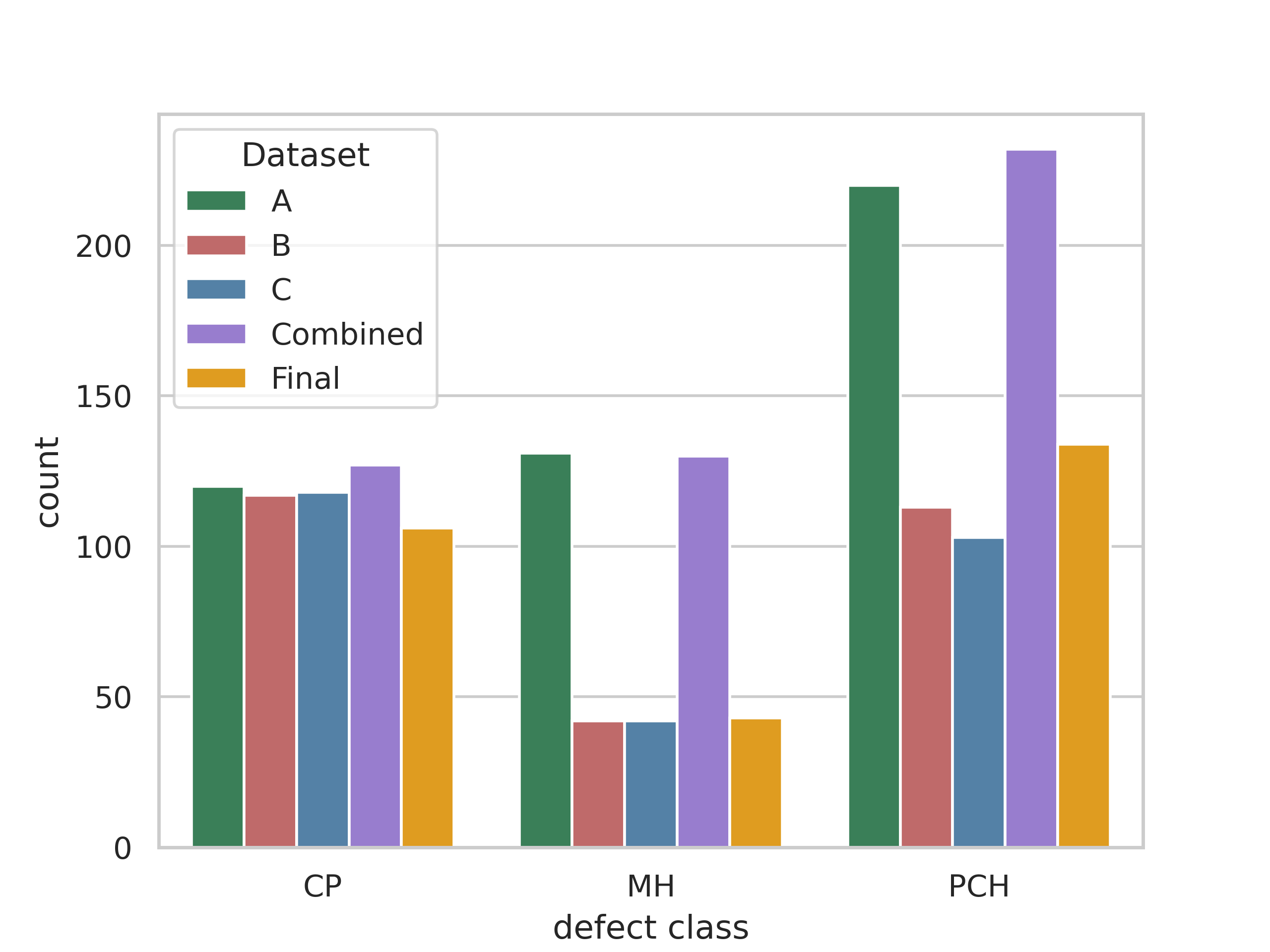} 
    \caption{All images except for MD images}
    \label{subfig:label_bar_graphs_no_md}
        \end{subfigure}
\caption{Label counts for each labeled dataset.}
\label{fig:label_bar_graphs}
\end{figure}

Figure \ref{fig:consensus_label_difference_examples} shows corresponding combined and final dataset labels for the examples shown in Figure \ref{fig:label_difference_examples}. It can be seen that the combined dataset contains far more labels than the final dataset. This is confirmed in Figure \ref{subfig:label_bar_graphs_all} which shows label count statistics for all images in each labeled dataset. Figure \ref{subfig:label_bar_graphs_no_md} shows the counts for all images except for MD images. The differences in label counts between datasets in Figure \ref{subfig:label_bar_graphs_all} are greater than those in Figure \ref{subfig:label_bar_graphs_no_md} which shows that most differences in labeling happen in MD images. We investigate the effects of training and evaluating with/without MD images, which have the highest likelihood to have erroneous labels due to the disagreement between labelers, in the next subsection.

\subsection{YOLOv8 defect detection performance}\label{subsect:model_performance}

\begin{figure}
    \centering
    \includegraphics[width=\linewidth]{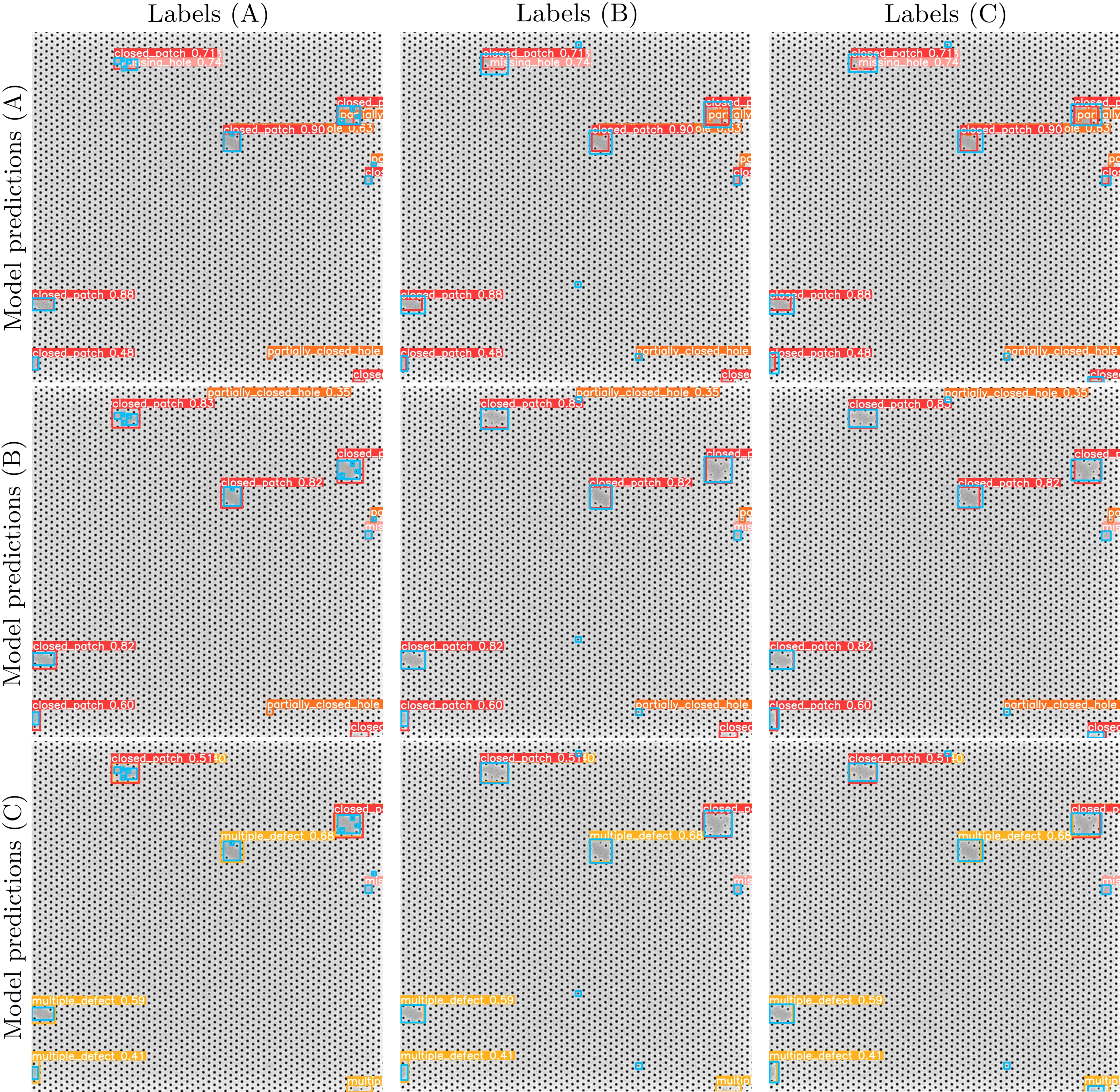}
\caption{Prediction examples on a test image for YOLOv8m (medium) models trained on the labels from labelers A, B, and C (top row, middle row, and bottom row, respectively). Displayed predictions include confidence scores. Each prediction example is overlayed with labels of the test image from labelers A, B, and C (left column, middle column, and right column, respectively). The labels are shown in blue without class information to improve visibility. Best viewed digitally and zoomed-in for the best visibility of the prediction and label details.}
\label{fig:pred_labelers}
\end{figure}

First, we investigate the performance of YOLOv8 models trained and tested only on the individual labeler datasets. Table \ref{tab:ind_labaler_results} shows the results of all combinations of train and test datasets from the three labelers. These results show that models trained and evaluated on data from the same labeler (A or B, not C due to the bad classification of MD labels) perform relatively well. Labeler C is an exception for this due to the model's classification confusion due to the erroneous MD labels. This behavior can be seen in the last row of Figure \ref{fig:pred_labelers}. The results from models trained on data from one labeler and tested on another labeler are significantly worse (again, when ignoring effects from MD labels). This shows that differences in labeling have a significant impact on model performance. The biggest differences are in the cross-labeler results involving labeler A since their label count was far greater than labelers B and C (see Figure \ref{subfig:label_bar_graphs_all}) which led to many false positive predictions compared to labeler B or C's test labels. This behavior can also be seen in Figure \ref{fig:pred_labelers}.


\begin{table}[h]
    \centering
    \caption{Per-class AP and mAP results for models trained and tested on datasets labeled by individual labelers. These results are the average of all YOLOv8 size variants. The first three rows of results use models that were trained and validated using the labels of one labeler and tested on the labels of the same labeler. The following rows use labels from different labelers for training and testing.}
    \label{tab:ind_labaler_results}
    \begin{tabular}{|c|c||c|c|c|c|c|}
        \hline
        \multicolumn{2}{|c|}{\textbf{Labeler}}  & \multicolumn{5}{c|}{\textbf{AP (@0.5 IoU)}} \\
        \cline{1-2} \cline{3-7}
         \textbf{train} & \textbf{test} & \textbf{CP} & \textbf{MH} & \textbf{PCH} & \textbf{MD} & \textbf{All (mAP)} \\
        \hline\hline
        \multirow{1}{*}{A} & \multirow{1}{*}{A} 
     & 0.883 & 0.873 & 0.769 & -- & 0.842 \\ \hline
        \multirow{1}{*}{B} & \multirow{1}{*}{B} 
     & 0.986 & 0.968 & 0.765 & -- & 0.906 \\ \hline
        C & C & 0.421 & 0.716 & 0.213 &  0.576 & 0.482 \\
        \hline\hline
        \multirow{2}{*}{A} & \multirow{1}{*}{B} 
     & 0.632 & 0.072 & 0.020 & -- & 0.260 \\ 
        & C & 0.396 & 0.052 & 0.231 & 0.0 & 0.226\\
        \hline
        \multirow{2}{*}{B} & \multirow{1}{*}{A} 
         & 0.504 & 0.096 & 0.020 & -- & 0.207 \\ 
        & C & 0.975 & 0.796 & 0.186 & 0.0 & 0.652 \\
        \hline
        \multirow{2}{*}{C} & A & 0.418 & 0.079 & 0.297 & -- & 0.265 \\
         & B & 0.975 & 0.804 & 0.195 & -- & 0.658 \\
        \hline
    \end{tabular}
\end{table}

Second, we investigate the performance of models trained and/or tested on the combined and final datasets. Table \ref{tab:final_test_results} shows the test results for all of these experiments. Similar to Table \ref{tab:ind_labaler_results}, we find that the best results come from models trained and evaluated on the same labels. The models trained on the final training data achieved the best result on the target test data, the test final dataset, of 0.919 mAP. This is the third-best overall mAP score achieved. The best overall mAP of 0.944 is achieved by the models trained on the final dataset without MD images and tested on the same data distribution. This suggests that our labeling combination and processing method is more coherent and complete than relying on individual labelers. The MD images are the most challenging which explains why the test results were better when the models are tested without them. Interestingly, however, models trained on the final dataset without MD images outperformed models trained with MD images for the CP and MH defect classes when tested using MD images (see red values in Figure \ref{tab:final_test_results}). This suggests that the final labels for the MD images are still a bit noisy. On the other hand, the improved performance of the PCH class when trained on MD images suggests that the optimal training dataset for the models should have a balance between label quality and quantity.

\begin{figure}
    \centering
\includegraphics[width=0.7\linewidth]{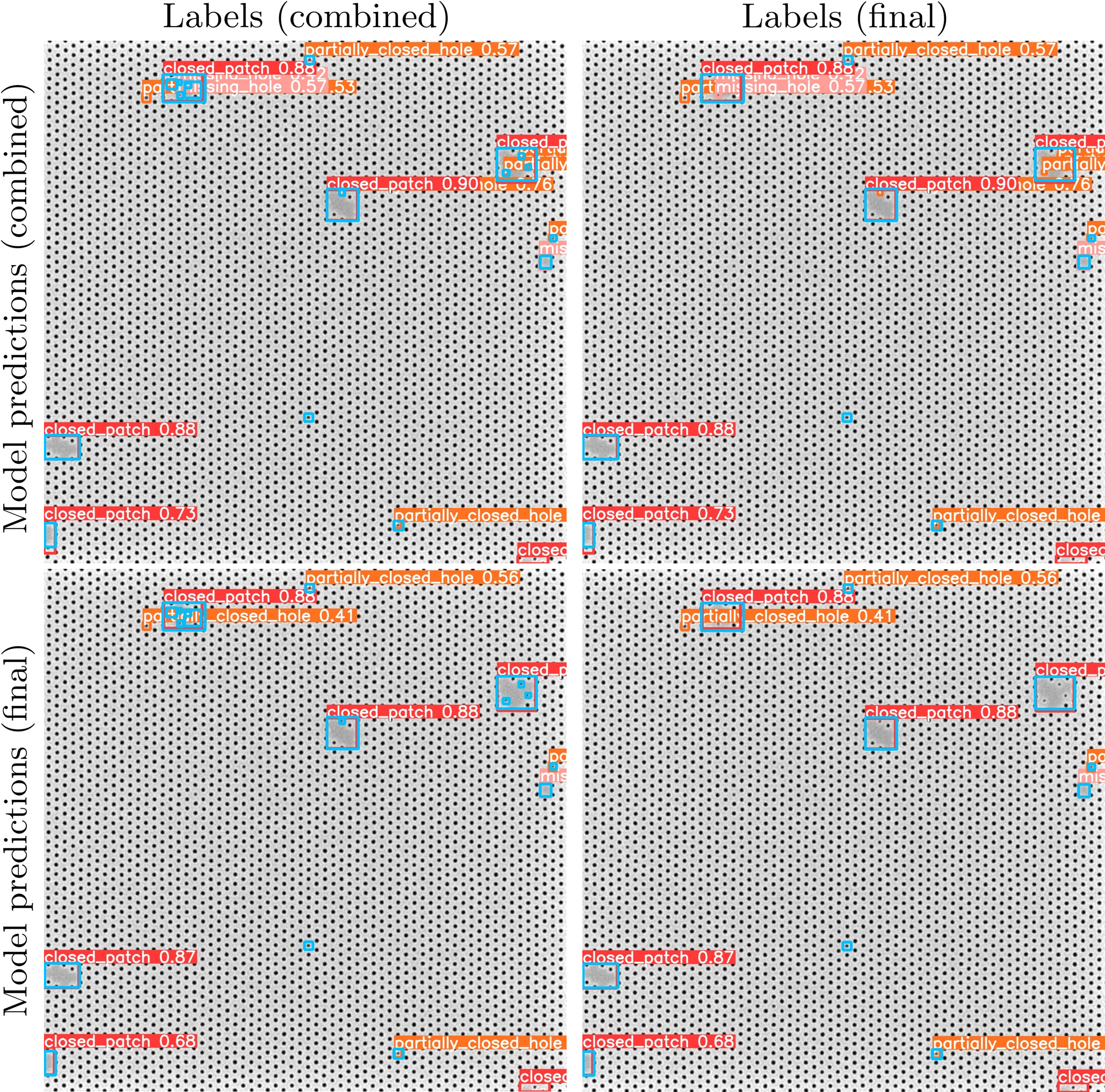}
\caption{Prediction examples on a test image for YOLOv8m (medium) models trained on the labels from the combined and final datasets (top row and bottom row, respectively). Displayed predictions include confidence scores. Each prediction example is overlayed with labels of the test image from the combined and final datasets (left column and right column, respectively). The labels are shown in blue without class information to improve visibility. Best viewed digitally and zoomed-in for the best visibility of the prediction and label details.}
\label{fig:pred_combined_and_final}
\end{figure}

We find that the performance of PCH detection, the most challenging defect class overall, improves most when trained on final dataset labels compared to training on individual labeler data. We also find that the performance of PCH detection increases significantly when no MD images are considered for testing. This could be indicative of false positive PCH labels in the final dataset (especially in MD images) due to the somewhat ambiguous characteristic of PCH defects. An example of such an erroneous label could be the PCH label just below the center of the input image from Figure \ref{fig:pred_labelers} where only labeler B, potentially erroneously, labeled this defect. Figure \ref{fig:pred_combined_and_final} shows that models trained on the combined or final datasets did not predict this defect at all. 

\begin{table}[h]
    \centering
    \caption{Per-class AP and mAP results for models trained and/or tested on the combined and/or final datasets. These results are the average of all YOLOv8 size variants. Post-processing refers to applying the ``expert-aware post-processing'' rules from Figure \ref{fig:consensus_algo_diagram} to model predictions before comparison with test labels. \textcolor{blue}{Blue} and \textcolor{red}{red} values are the best test values for combined and final data, respectively.}
    \label{tab:final_test_results}
    \begin{tabular}{|c|c||c|c|c|c|}
        \hline
        \multicolumn{2}{|c|}{\textbf{Dataset}} & \multicolumn{4}{c|}{\textbf{AP (@0.5 IoU)}} \\
        \hline
         \textbf{train} & \textbf{test} & \textbf{CP} & \textbf{MH} & \textbf{PCH} & \textbf{All (mAP)} \\
        \hline\hline
        \multirow{2}{*}{A} & Combined  & 0.856 & 0.872 & 0.655 & 0.794 \\ 
         & Final  & 0.810 & 0.072 & 0.268 & 0.383 \\ \hline
        \multirow{2}{*}{B} & Combined  & 0.892 & 0.190 & 0.235 & 0.438 \\ 
         & Final  & 0.976 & 0.968 & 0.383 & 0.775 \\ \hline
        \multirow{2}{*}{C} & Combined  & 0.911 & 0.125 & 0.290 & 0.441 \\ 
         & Final  & 0.931 & 0.792 & 0.460 & 0.728 \\ \hline
        \multirow{4}{*}{Combined} & \multirow{1}{*}{Combined}  & \textcolor{blue}{0.945} & \textcolor{blue}{0.904} & \textcolor{blue}{0.795} & \textcolor{blue}{0.881} \\
         & \multirow{1}{*}{Final}  & 0.944  & 0.222 & 0.550 & 0.572 \\
         & \multirow{1}{*}{Final (+ post-processing)}  & 0.974 & 0.810 & 0.578 & 0.787 \\
         & \multirow{1}{*}{Final (no MDs + post-processing)}  & 0.986 & 0.911 & 0.719 & 0.872 \\
        \hline
        \multirow{2}{*}{Final} & \multirow{1}{*}{Final}  & 0.977 & 0.984 & \textcolor{red}{0.797} & \textcolor{red}{0.919} \\
         & \multirow{1}{*}{Final (no MDs)}  & 0.952 & 0.980 & 0.844 & 0.925 \\
        \hline
        \multirow{2}{*}{Final (no MDs)} & \multirow{1}{*}{Final}  & \textcolor{red}{0.992} & \textcolor{red}{0.995} & 0.695 & 0.892 \\
         & \multirow{1}{*}{Final (no MDs)}  & 0.995 & 0.984 & 0.853 & 0.944 \\
        \hline
    \end{tabular}
\end{table}

The models trained on labeler A's dataset perform decently well on the combined test dataset and poorly on the final dataset. The opposite is true for models trained on labeler B and C's datasets which perform better on the final dataset than the combined dataset. This is because many of the labels in the combined dataset come exclusively from labeler A.

Applying the same post-processing steps used to create the final dataset to the predictions of the model trained on the combined dataset significantly improves the performance when tested on the final test dataset. However, this improved performance is still noticeably lower than the results obtained by the models trained on the final dataset itself. Post-prediction-processing in this manner could still be useful in a dynamic environment where new post-processing rules are regularly introduced and re-training of the model is not wanted.

\section{Limitations \& Future Work}\label{sect:future_work}
In our proposed labeling combination method, we took a liberal approach to combining labels in the sense that we assume all labels are valid labels for defects. This minimizes false negatives. However, this means that any false positive labels from individual labelers will also be included in the combined dataset. This can be an issue for defects that can sometimes be similar to the background pattern such as PCH as discussed in the previous section. Future work could experiment with more conservative label combination approaches or automatic erroneous label detection and removal methods.

\begin{figure}
    \centering
    \begin{subfigure}[b]{0.63\linewidth}
        \includegraphics[width=\textwidth]{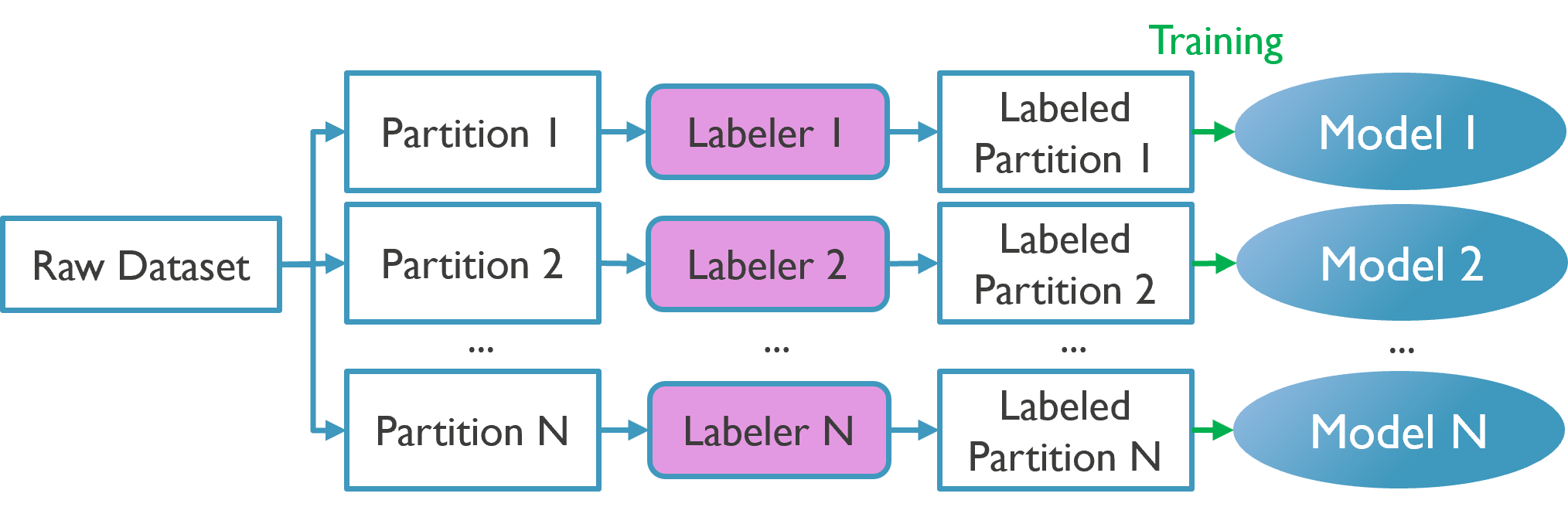}
        \caption{Per partition labeling and model training. Note each model is the same architecture but is trained on different partitions.}
        \label{fig:fw_a}
    \end{subfigure}\hfill
    \begin{subfigure}[b]{0.33\linewidth}       \includegraphics[width=\textwidth]{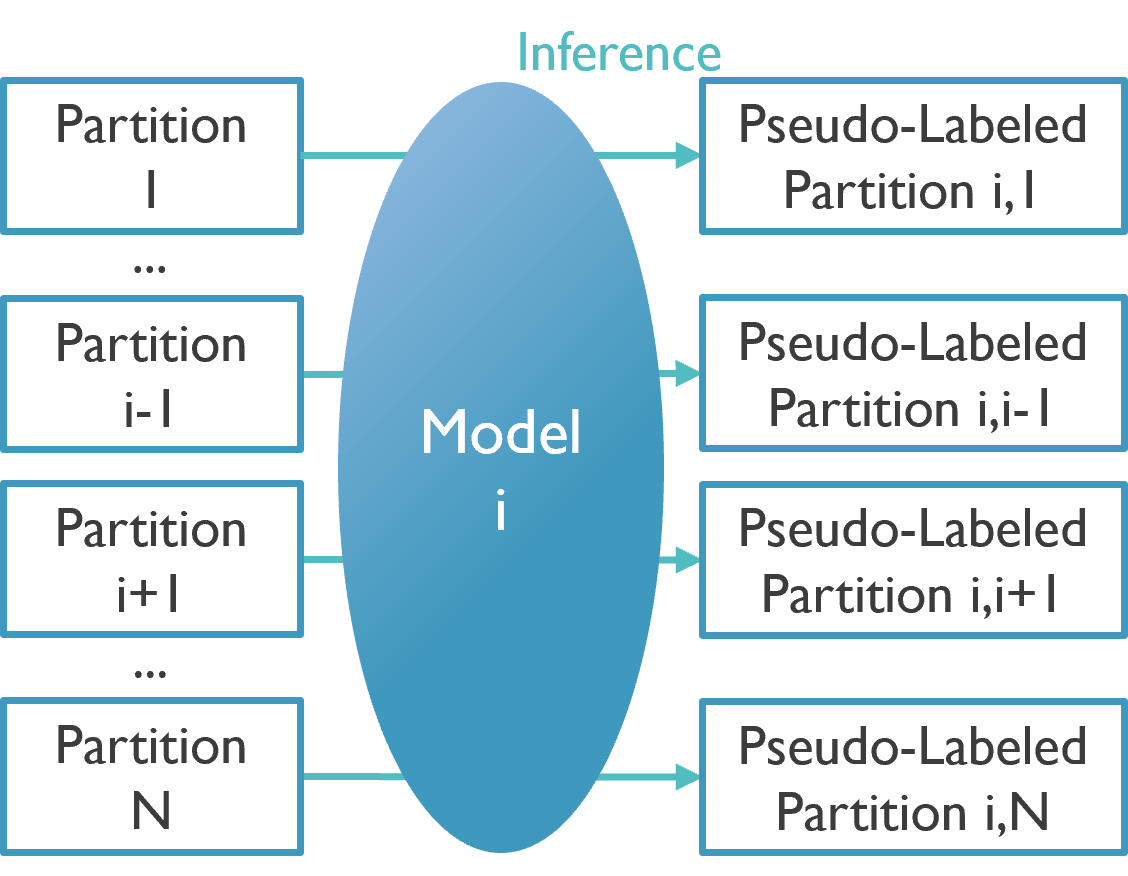}
        \caption{Prediction of pseudo-labels by a model trained on labeled partition i.}
        \label{fig:fw_b}
    \end{subfigure}
    \caption{Suggested method for semi-automated labeling of large datasets in future work.}
    \label{fig:fw_all}
\end{figure}

Another limiting assumption we make is that each image is labeled by multiple labelers. This is inefficient in the sense that a lot of duplicate work is performed, especially for labeling ``obvious'' defects. This could be improved in future work given access to larger datasets by (i) having different partitions of the dataset labeled only once by different labelers (see Figure \ref{fig:fw_a}), (ii) training separate models on each partition (see also Figure \ref{fig:fw_a}), (iii) using these models to predict pseudo-labels on the other image partitions it was not trained on (see Figure \ref{fig:fw_b}), and (iv) applying a label combination algorithm on all the manual- and pseudo-labels for each image in the dataset. This allows larger datasets to be labeled using similar amounts of manual labeling effort while benefitting from the coherency and completeness of our proposed label combination method. Future work should take care to identify major labeling errors (such as the MD misclassifications by labeler C which led to poor performance even when tested on their own dataset as shown in Table \ref{tab:ind_labaler_results}) and ensure that each labeler labels enough images to train a model that can reliably predict pseudo-labels similar to the labeling behavior the model was trained on.

\section{Conclusion}\label{sect:conclusion}
ML has the potential to reliably detect defects for a wide variety of patterns. This includes exploratory patterns such as HEXCH DSA patterns. However, the performance of supervised ML-based defect detection models is dependent on the quality of the labels of its training data. In this work, we proposed a method of combining labels from multiple human labelers to create a more coherent and complete final dataset. This final dataset best reflected the labels expected by a DSA expert without requiring exhaustive manual inspection and label correction from the DSA expert. Experimental results showed that state-of-the-art YOLOv8 models can achieve high detection precisions of more than 0.9 mAP when trained and tested on the final dataset. Finally, we suggested future work continue in the direction of data-centric ML by exploring automatic defect label removal and pseudo-label prediction.

\bibliography{report}
\bibliographystyle{spiebib} 

\end{document}